\documentclass[
]{ceurart}

\usepackage{listings}
\usepackage[noabbrev]{cleveref}
\usepackage{tikz}
\usepackage{siunitx}
\begin{document}

\copyrightyear{2024}
\copyrightclause{Copyright for this paper by its authors.
	Use permitted under Creative Commons License Attribution 4.0
	International (CC BY 4.0).}

\conference{4th Italian Workshop on Artificial Intelligence and Applications for Business and Industries --- AIABI | co-located with AI*IA 2024}

\title{Flowcean --- Model Learning for Cyber-Physical Systems}

\author[1]{Maximilian Schmidt}[%
	orcid=0009-0005-4532-7669,
	email=maximilian.schmidt@tuhh.de,
]
\author[1]{Swantje Plambeck}[%
	orcid=0000-0002-4875-5280,
	email=swantje.plambeck@tuhh.de,
]
\author[1]{Markus Knitt}[%
	orcid=0000-0002-4051-2268,
	email=markus.knitt@tuhh.de,
]
\author[1]{Hendrik Rose}[%
	orcid=0000-0002-7850-5071,
	email=hendrik.rose@tuhh.de,
]
\author[1]{Goerschwin Fey}[%
	orcid=0000-0001-6433-6265,
	email=goerschwin.fey@tuhh.de,
]
\address[1]{Hamburg University of Technology, Hamburg, Germany}
\author[2]{Jan Christian Wieck}[%
	orcid=0000-0003-0330-2939,
	email=jan.christian.wieck@cml.fraunhofer.de,
]
\address[2]{Fraunhofer --- Center for Maritime Logistics and Services CML, Hamburg, Germany}

\author[3]{Stephan Balduin}[%
	orcid=0000-0002-2018-1078,
	email=stephan.balduin@offis.de,
]
\address[3]{OFFIS --- Institute for Information Technology, Oldenburg, Germany}

\begin{abstract}
	Effective models of Cyber-Physical Systems (CPS) are crucial for their design and operation.
	Constructing such models is difficult and time consuming due to the inherent complexity of CPS\@.
	As a result, data-driven model generation using machine learning methods is gaining popularity.
	In this paper, we present Flowcean, a novel framework designed to automate the generation of models through data-driven learning that focuses on modularity and usability.
	By offering various learning strategies, data processing methods, and evaluation metrics, our framework provides a comprehensive solution, tailored to CPS scenarios.
	Flowcean facilitates the integration of diverse learning libraries and tools within a modular and flexible architecture, ensuring adaptability to a wide range of modeling tasks.
	This streamlines the process of model generation and evaluation, making it more efficient and accessible.
\end{abstract}


\maketitle

\section{Introduction}\label{sec:introduction}

Cyber-Physical Systems (CPS) are crucial in many industrial sectors, including energy, mobility, and logistics~\cite{yaacoubCyberphysicalSystemsSecurity2020}.
These systems integrate physical components such as sensors and actuators with digital logic~\cite{yohanandhanCyberPhysicalPowerSystem2020}.
As a result, CPS pose a range of interdisciplinary challenges; ensuring their reliability and security are only two.
In response to these challenges, we introduce a novel machine learning framework for CPS\@.
The following sections outline the motivation for this development, provide a relevant background from related work, and highlight the main contributions of this paper.

\subsection{Motivation}\label{sec:motivation}

CPSs represent a broad category of systems, often encompassing expensive components or critical infrastructure, such as specialized hardware or large-scale systems like the electrical power grid.
Despite their diversity, CPSs share common characteristics: they are highly complex, often not fully understood, and typically unsuitable for direct use in development processes.
Consequently, CPS models are essential for tasks such as design, verification, and testing, as they provide a clearer understanding of system behavior.

The diversity of CPS domains brings additional modeling challenges.
Conventional physics-based modeling approaches require significant manual effort, extensive domain knowledge, and computational expertise.
All of which scale with the complexity of the system~\cite{derlerModelingCyberPhysical2012a}.
Moreover, the absence of standardized modeling frameworks, along with the wide range of CPS applications, means that each system has a unique design and configuration.
This variation complicates the reuse of models and learning processes in different CPSs, making manual modeling time consuming and difficult~\cite{knittAutomaticGenerationModels2023}.
For some systems, manual modeling may not even be feasible due to their complexity or the lack of necessary domain expertise.

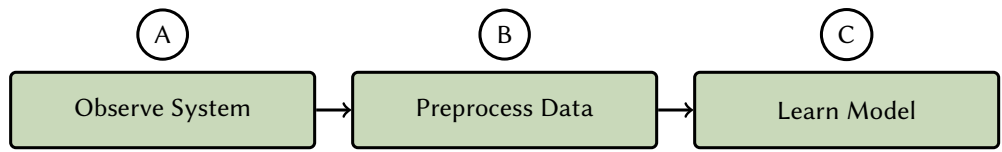
\begin{figure}[htbp]
	\centering
	\tikzset{
  every picture/.style={line width=0.4mm},
}
\begin{tikzpicture}[every node/.style={align=center}]

    \definecolor{mycolor}{rgb}{0.3, 0.5, 0.1}

    \tikzstyle{box} = [draw, rounded corners=2pt, minimum width=4cm, minimum height=1cm, fill=mycolor, fill opacity=0.3, text opacity=1]
    
    \node[box] (observe) at (0, 0) {Observe System};
    \node[box] (preprocess) at (4.5, 0) {Preprocess Data};
    \node[box] (learn) at (9, 0) {Learn Model};

    \node[draw, circle, minimum size=0.5cm] at (0, 1.0) {A};
    \node[draw, circle, minimum size=0.5cm] at (4.5, 1.0) {B};
    \node[draw, circle, minimum size=0.5cm] at (9, 1.0) {C};

    \draw[->] (observe) -- (preprocess);
    \draw[->] (preprocess) -- (learn);

\end{tikzpicture}
	\caption{Abstract workflow of data-driven modeling}\label{fig:data-driven-modeling}
\end{figure}

Given these challenges, data-driven modeling has emerged as a promising alternative, as it reduces the need for manual effort and specialized knowledge.
Data-driven modeling involves the automatic generation of models from data collected from the system.
\Cref{fig:data-driven-modeling} provides a high-level overview of this process.

The process begins with system observation (A), where data is collected from various sources, such as precollected datasets, online simulations, or interactions with real-world CPS\@.
The data is then pre-processed (B) through tasks such as standardization or feature selection, which determine how the learning algorithm interprets the information and establishes an abstraction between the system and its model.
In the model learning phase (C), a learning algorithm processes the data to generate a model that captures the behavior of the system.

Since each CPS is unique, the data-driven learning pipeline must be customized to the specific characteristics of the system.
As a result, these pipelines often produce highly specialized solutions, tailored to individual CPS\@.

\subsection{Background}\label{sec:background}

Machine learning frameworks such as PyTorch and TensorFlow have made significant strides in modeling the behavior of CPS\@.
The flexibility of these frameworks has been effectively used in tasks such as anomaly detection within digital twins~\cite{xuDigitalTwinbasedAnomaly2021} and the development of learning-based adversarial agents in CPS environments~\cite{khanapuriLearningBasedAdversarialAgent2019}.
These frameworks support the processing of large-scale datasets and the design of complex neural network architectures, making them suitable for a wide range of CPS applications~\cite{fengTimeSeriesAnomaly2021}.

For time-discrete CPS, traditional approaches like automata learning are commonly used to model system behavior.
Automata can be learned through active system interaction~\cite{urbatAutomataLearningAlgebraic2020a, steffenIntroductionActiveAutomata2011a} or passive observation of system behavior~\cite{maierOnlinePassiveLearning2014}.
These approaches are particularly effective for systems where the state space is discrete and the dynamics are governed by rules or transitions between states.
For nondeterministic systems, decision trees have also been employed as a simple yet powerful approach to model behavior~\cite{plambeckViabilityDecisionTrees2022, cuiAnalogCircuitsFault2016}.
Although effective, these techniques often depend on domain-specific algorithms and typically require custom implementations tailored to each new application.

Libraries such as Keras and scikit-learn have been popularizing machine learning by providing accessible and efficient implementations of common algorithms.
Keras, known for its user-friendly interface, enables rapid prototyping and experimentation with neural networks.
scikit-learn offers a wide array of tools for data mining and analysis, including regression, classification, and clustering algorithms, which have been applied in CPS for tasks such as fault detection~\cite{meleshkoMachineLearningBased2020} and predictive maintenance~\cite{kumarsharmaDataDrivenPredictive2022}.
Although these frameworks excel at general-purpose machine learning tasks, they are less equipped to meet the specific demands of CPS, such as real-time data processing or the integration of heterogeneous data sources.

\subsection{Contribution}\label{sec:contribution}

Although existing machine learning frameworks are powerful, they are typically designed for specific learning strategies, such as passive learning on static datasets or reinforcement learning in controlled environments.
This specialization limits their flexibility in accommodating different approaches or generalizing across various CPS applications.
Adapting the same system for different learning strategies typically requires modifying the learning library and restructuring the data pipelines.
For example, a framework optimized for passive learning may not efficiently handle incremental learning scenarios, where models must be continuously updated as new data become available.
Similarly, deep learning frameworks may not support automata-based approaches or decision trees, which are better suited for certain types of CPS\@.
Due to the complexity of CPS behavior, it is often difficult to predict the optimal approach and framework before modeling begins, leading to duplicated effort if a change in framework becomes necessary mid-process.

To address these limitations and challenges, we introduce \emph{Flowcean}, a modular framework designed specifically to generate CPS models.
Flowcean offers several key contributions to existing machine learning frameworks by addressing the unique needs of CPS modeling:

\begin{itemize}
	\item
	      \emph{Re-usability of learning processes and components:}
	      Flowcean provides a flexible learning pipeline that defines reusable paradigms and interfaces, enabling modules to be applied across different CPS applications without significant reconfiguration.
	\item
	      \emph{Integration and combination of learning strategies:}
	      Flowcean allows the seamless integration of multiple learning strategies, ranging from passive to incremental and active approaches, within a unified framework.
	      By interfacing with existing learning libraries, it broadens applicability across diverse domains.
	\item
	      \emph{Common CPS data formats:}
	      Flowcean supports a wide range of CPS data formats, including tabular CSV, time series data, ROS bag files, and direct connections to real-world systems via gRPC, facilitating easy data integration for CPS modeling.
	\item
	      \emph{Application scenarios:}
	      Flowcean covers a wide range of tasks with a particular focus on CPS, aligning closely with the unique diversity in both cyber and physical parts of CPS, e.g., system prediction or monitoring.
\end{itemize}

The remainder of this paper is structured as follows.
In \Cref{sec:concepts}, we introduce the core concepts underlying Flowcean and propose a novel approach to modularize learning pipelines.
In \Cref{sec:flowcean}, we describe the architecture of Flowcean in detail.
A practical application is demonstrated through a case study in \Cref{sec:case_study}, and we conclude with a summary of our findings and future work directions in \Cref{sec:conclusion}.

Flowcean, along with the case studies, is publicly available under the BSD 3-Clause license on GitHub\footnote{\url{https://github.com/flowcean/flowcean.}}.

\section{Concept}\label{sec:concepts}

\begin{figure}[htbp]
	\centering
	\tikzset{
  every picture/.style={line width=0.4mm},
}
\usetikzlibrary{shadings}
\begin{tikzpicture}[every node/.style={align=center}]

    \definecolor{mygreen}{rgb}{0.3, 0.3, 0.3}
    \definecolor{myblue}{rgb}{0.2, 0.3, 1.0}
    \definecolor{myred}{rgb}{0.8, 0.1, 0.1}

    \tikzstyle{box} = [draw, rounded corners=2pt, minimum width=2cm, minimum height=1cm, fill opacity=0.3, text opacity=1]
    
    \node[box, fill=mygreen] (rectTa) at (0.75, 0) {CPS Modeling Task};

    \node[box, fill=myblue] (rectLS) at (-1.8, -1.5) {Learning Strategy};
    \node[box, fill=myred]  (rectES) at (3.55, -1.5) {Evaluation Strategy};

    \node[box, shading = axis, left color=myblue, right color=myred, minimum width=2.2cm] (rectE1) at (-3.5, -3) {A \\ Environment};
    \node[box, fill=myblue, minimum width=1.7cm] (rectT)  at (-1.4, -3) {B \\ Transform};
    \node[box, fill=myblue, minimum width=1.3cm] (rectL)  at (0.3, -3)  {C \\ Learner};
    
    \node[box, shading = axis, left color=myblue, right color=myred, minimum width=2.2cm] (rectE2) at (2.2, -3)  {D \\ Environment};
    \node[box, fill=myred, minimum width=1.1cm] (rectMo) at (4.0, -3)  {E \\ Model};
    \node[box, fill=myred, minimum width=1.3cm] (rectMe) at (5.4, -3)  {F \\ Metric};

    \draw[->] (rectL) -- (rectLS);
    \draw[->] (rectT) -- (rectLS);
    \draw[->] (rectE1) -- (rectLS);
    \draw[->] (rectE2) -- (rectES);
    \draw[->] (rectMo) -- (rectES);
    \draw[->] (rectMe) -- (rectES);
    
    \draw[->] (rectLS) -- (rectTa);
    \draw[->] (rectES) -- (rectTa);

\end{tikzpicture}
	\caption{Component-view of the concepts, showing the three steps of data-driven modeling (A to C) extended by an evaluation (D to F)}\label{fig:overview}
\end{figure}
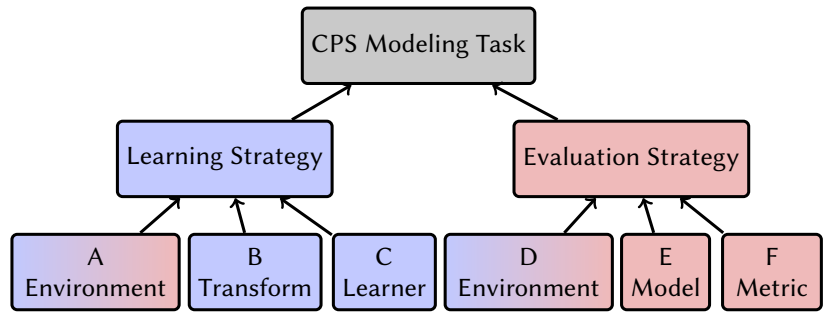

We introduce a modular pipeline for data-driven modeling of CPS\@.
\Cref{fig:overview} shows the architectural relation between different concepts, as well as the six steps (A to F) of model generation and evaluation.
The modeling task is partitioned into learning and evaluation.

The initial step in the workflow involves the acquisition of data from the system.
We refer to a data source as an \emph{environment (A, D)}, which encompasses precollected data sets, online system simulations, or real-world CPS, providing a generalized interface to both the learning and the evaluation strategy.

Data from the environment may undergo preprocessing through \emph{transforms (B)}.
Following this, the transformed data is fed to a \emph{learner (C)} to generate a model.
The composition of the environment, transforms and the learner constitutes a \emph{learning strategy}.
In the \emph{evaluation strategy}, a second environment (D) is used to evaluate the generated \emph{model (E)} on unseen data and assess its performance using \emph{metrics (F)}.

With this conceptual view, we provide a generalized abstraction while capturing existing machine learning and modeling paradigms.
By defining interfaces, we craft re-usable modules that allow the construction of individual pipelines.
The following sections delve into the underlying concepts of all components outlined in \Cref{fig:overview} and show further connections to established machine learning paradigms.

\subsection{Environments}\label{sec:environments}

Environments abstract the data source and provide a unified entry point to learning strategies.
Generally, we differentiate three ways of data accessibility: a precollected data set, a stream of observations of a CPS, or an active interaction with a system.
We therefore define three types of environment: Offline, Incremental, and Active.

\begin{itemize}
	\item
	      An \emph{Offline Environment} provides an interface to load a data set.
	      Data is precollected and saved in a file or a database.
	      The environment is observed once and provides the data as a single batch.
	\item
	      An \emph{Incremental Environment} provides a stream of observations.
	      That is, the environment continuously provides data to the learning strategy.
	      This may be a real-time stream of sensor data from a simulation or a batchwise replay of pre-collected data.
	\item
	      An \emph{Active Environment} provides an interface to interact with the system.
	      Data is generated on demand by actively engaging with the system.
	      This may involve a simulation or a real system.
\end{itemize}

Each of the three environments has a tight coupling with the way data is collected and made available to the learning process.
Consequently, each environment only serves compatible learning or evaluation strategies.

\subsection{Transforms}\label{sec:transforms}

The preparation of data for machine learning tasks is crucial for the performance of a learned model~\cite{fanReviewDataPreprocessing2021}.
We introduce transforms as all types of operations performed on data samples or data sets, facilitating tasks such as pre-processing, feature engineering, and data augmentation.
All of these operations involve the same fundamental concept of transforming data, incorporating tasks such as normalization, dimensionality reduction, one-hot encoding, and others.

The data preparation process often involves the use of multiple transforms.
By chaining them, we create a pipeline that prepares the data for modeling and defines the abstraction between the original system and the generated model.
Transforms act as a bridge between the system and the learner, especially in cases where a continuous system is abstracted into a discrete model.

When designing a learning pipeline, it is essential to determine whether a transformation is specific to the data or to the learner, as this dictates where the transform should be applied within the pipeline.
One of Flowcean's main features here is to maintain modularity and flexibility, each transform being treated as a separate component that can be attached to either the environment or passed to the learning strategy.
This approach ensures that the pipeline remains functional and adaptable, allowing for the seamless integration of different environments or learning strategies without the need for extensive reconfiguration.

\subsection{Learning Strategies}\label{sec:learning_strategies}

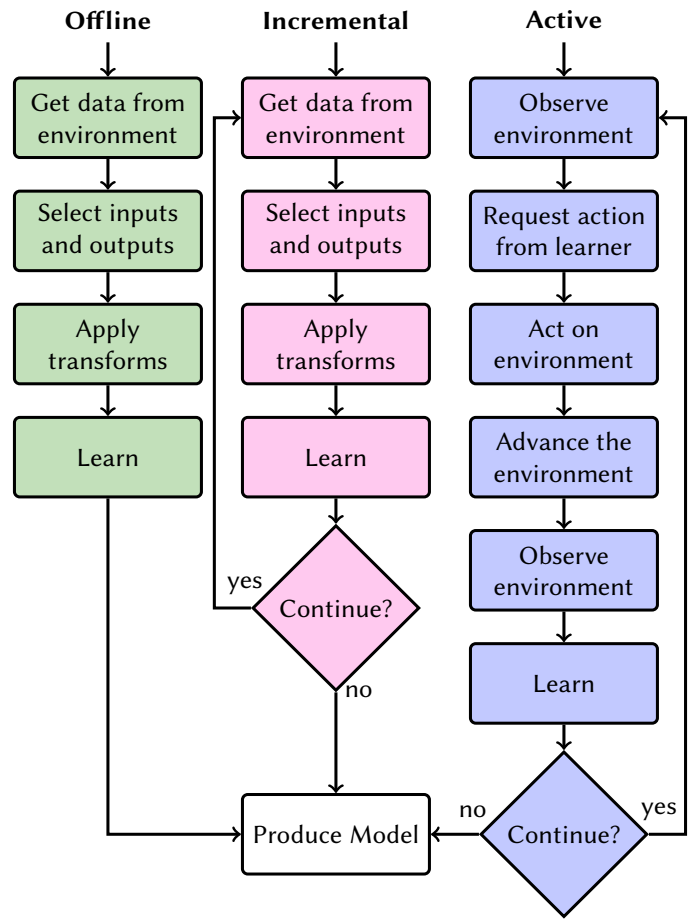
\begin{figure}[htbp]
	\centering
	\usetikzlibrary{shapes.geometric}
\usetikzlibrary{arrows}
\tikzset{
  every picture/.style={line width=0.4mm},
}
\begin{tikzpicture}[every node/.style={align=center}, scale=1]

    \definecolor{incrementalcolor}{rgb}{1.0, 0.3, 0.8}
    \definecolor{offlinecolor}{rgb}{0.2, 0.6, 0.1}
    \definecolor{activecolor}{rgb}{0.2, 0.3, 1.0}

    \tikzstyle{textBox} = [draw, rounded corners=2pt, minimum width=7em, minimum height=3em, fill opacity=0.3, text opacity=1]
    \tikzstyle{decisionBox} = [draw, diamond, aspect=1, minimum width=4em, minimum height=4em, fill opacity=0.3, text opacity=1]
    
    \node at (0,-0.2) {\textbf{Incremental}};
    \node[textBox, fill=incrementalcolor] (incremental_0) at (0,-1.5) {Get data from\\ environment};
    \node[textBox, fill=incrementalcolor] (incremental_1) at (0,-3) {Select inputs\\ and outputs};
    \node[textBox, fill=incrementalcolor] (incremental_2) at (0,-4.5) {Apply\\ transforms};
    \node[textBox, fill=incrementalcolor] (incremental_3) at (0,-6) {Learn};
    \node[decisionBox, fill=incrementalcolor] (continue) at (0,-8) {Continue?};
    \node[textBox] (return) at (0,-11) {Produce Model};

    \draw[->] (0, -0.5) -- (incremental_0);
    \draw[->] (incremental_0) -- (incremental_1);
    \draw[->] (incremental_1) -- (incremental_2);
    \draw[->] (incremental_2) -- (incremental_3);
    \draw[->] (incremental_3) -- (continue);
    \draw[->] (continue) -- (return);
    \draw[->] (continue) -| (-1.6, -8) |- (incremental_0);
    \node at (-1.2, -7.7) {yes};
    \node at (0.3, -9.1) {no};

    \node at (-3,-0.2) {\textbf{Offline}};
    \node[textBox, fill=offlinecolor] (offline_0) at (-3,-1.5) {Get data from\\ environment};
    \node[textBox, fill=offlinecolor] (offline_1) at (-3,-3) {Select inputs\\ and outputs};
    \node[textBox, fill=offlinecolor] (offline_2) at (-3,-4.5) {Apply\\ transforms};
    \node[textBox, fill=offlinecolor] (offline_3) at (-3,-6) {Learn};

    \draw[->] (-3, -0.5) -- (offline_0);
    \draw[->] (offline_0) -- (offline_1);
    \draw[->] (offline_1) -- (offline_2);
    \draw[->] (offline_2) -- (offline_3);
    \draw[->] (offline_3) |- (return);

    \node at (3,-0.2) {\textbf{Active}};
    \node[textBox, fill=activecolor] (active_0) at (3,-1.5) {Observe\\ environment};
    \node[textBox, fill=activecolor] (active_1) at (3,-3) {Request action\\ from learner};
    \node[textBox, fill=activecolor] (active_2) at (3,-4.5) {Act on\\ environment};
    \node[textBox, fill=activecolor] (active_3) at (3,-6) {Advance the\\ environment};
    \node[textBox, fill=activecolor] (active_4) at (3,-7.5) {Observe\\ environment};
    \node[textBox, fill=activecolor] (active_5) at (3,-9) {Learn};
    \node[decisionBox, fill=activecolor] (finished) at (3,-11) {Continue?};

    \draw[->] (3, -0.5) -- (active_0);
    \draw[->] (active_0) -- (active_1);
    \draw[->] (active_1) -- (active_2);
    \draw[->] (active_2) -- (active_3);
    \draw[->] (active_3) -- (active_4);
    \draw[->] (active_4) -- (active_5);
    \draw[->] (active_5) -- (finished);
    \draw[->] (finished) -- (return);
    \draw[->] (finished) -| (4.6, -11) |- (active_0);
    \node at (4.25, -10.7) {yes};
    \node at (1.8, -10.7) {no};

\end{tikzpicture}
	\caption{Flowchart for the variants of learning strategies}\label{fig:learning_strategies}
\end{figure}

Learning strategies integrate an environment, transforms, and a learner to generate a model.
They define the learning process and the model update mechanism by establishing a learning pipeline.
To our knowledge, Flowcean is the first framework to inherently combine three distinct variants of learning strategies: offline, incremental, and active learning.
\Cref{fig:learning_strategies} visualizes these learning strategies in a flow chart.

Offline learning strategies involve processing a fixed data set at once, in which the model parameters are updated collectively.
An environment provides the relevant data, and the learning strategy selects the respective features to represent the input and output of the system.
Supervised learning is an example of an offline learning strategy, where the model is trained on a pre-collected data set.

In contrast, incremental learning occurs by continuously updating the model as new data becomes available.
That is, an environment incrementally provides data to the learning strategy, and in each incremental step, the learner processes small batches of data, potentially even individual samples.
Here, the learner is passive, i.e., does not influence or interact with the data source.
The environment autonomously decides which samples to present to the learner, rendering the process non-interactive.
Learning from streaming data is, thus, an example of incremental learning.
This process is also known as online learning in the literature, although there is no fixed terminology on this distinction~\cite{ofjallOnlineLearningRobot2014}.

Active learning engages the learner in actively interacting with the environment.
In this scenario, the learning strategy requests an action from the learner, which in turn advances the environment.
The environment then provides data samples that are observed and provided to the learner.
This interactive process is exemplified in approaches such as reinforcement learning, where the decisions of a learner affect the selection of data or the simulation.

\subsection{Model}\label{sec:model}

A model is the product of a learning strategy.
While the learner is an algorithm that trains the model by learning from the data, the model represents the final abstraction of the system.
Flowcean distinguishes between the learner and the model to enhance modularity and segregate responsibilities.
This also allows the model to be used in application scenarios without reliance on the learning algorithm.
Once the learning strategy has finished training, it generates a model that represents the acquired knowledge that the learner has extracted from the data.
The model can subsequently be used to, e.g., process or predict new, unseen data.

\subsection{Evaluation Strategies}\label{sec:evaluation_strategies}

In the evaluation, the performance of a model is assessed.
This helps in making informed decisions about the deployment of models and allows for comparison of models generated by different learners.
As depicted in \Cref{fig:overview}, an evaluation strategy defines the evaluation composed of an environment, a model, and metrics.
A strategy evaluating the predictive performance of a model commonly follows the following steps:

\begin{enumerate}
	\item
	      Retrieve evaluation data from the environment
	\item
	      Predict using the learned model
	\item
	      Compare the model output with the system behavior using metrics
\end{enumerate}

Similarly to learning strategies, evaluation strategies depend on the type of application scenario and the model.
For instance, models generated for an interactive application might require different evaluation strategies than models generated for passive monitoring applications.

\section{Architecture of Flowcean}\label{sec:flowcean}

In this section, we cover the implementation details of Flowcean.
Using the concepts outlined in the preceding sections, Flowcean allows the integration of various learning libraries and tools.
This approach enables the usage of different models with identical interfaces, thus mitigating the effort to reimplement learning scenarios for similar applications.

\subsection{Modules}\label{sec:modules}

At the core of Flowcean lies a set of abstract classes specifying the interfaces of \Cref{sec:concepts} to serve as blueprints for the implementation of concrete modules.
This modular architecture enables the integration of learning libraries and tools, ensuring adaptability across various modeling tasks.

Alongside base classes, Flowcean provides concrete implementations for offline, incremental, and active environments, covering the three types of data sources for the learning strategies outlined in \Cref{fig:learning_strategies}.
Learning and evaluation strategies are functions accepting respective modules as arguments.
In addition to modules and strategies, Flowcean provides a utility for logging and loading data from common formats such as CSV or ROS bags, allowing quick adaptation to specific modeling tasks.

\subsection{State of Implementation}\label{sec:state_of_implementation}

\begin{table}[htbp]
	\centering
	\caption{Existing implementations of Flowcean modules}\label{tab:modules}
	\begin{tabular}{ll}
		\toprule
		\textbf{Module} & \textbf{Existing Implementations} \\
		\midrule
		Environment     &
		\begin{tabular}[c]{@{}l@{}}
			Loader for CSV, JSON, Parquet, YAML, ROS bag \\
			Interfaces to Polars and PyTorch data sets   \\
			ODE environment
		\end{tabular}        \\
		\midrule
		Transform       &
		\begin{tabular}[c]{@{}l@{}}
			Explode        \\
			Select         \\
			Sliding window \\
			Standardize
		\end{tabular}                          \\
		\midrule
		Learner         &
		\begin{tabular}[c]{@{}l@{}}
			External interface via gRPC      \\
			Neural network lightning learner \\
			Linear regression                \\
			Regression tree
		\end{tabular}                    \\
		\midrule
		Model           &
		\begin{tabular}[c]{@{}l@{}}
			PyTorch model      \\
			scikit-learn model \\
			gRPC model
		\end{tabular}                          \\
		\midrule
		Metric          &
		\begin{tabular}[c]{@{}l@{}}
			\textit{Regression:}                  \\
			\quad Max error, Mean absolute error, \\
			\quad Mean squared error, R2 score    \\
			\textit{Classification:}              \\
			\quad Accuracy, Precision, Recall, F-beta score
		\end{tabular}      \\
		\bottomrule
	\end{tabular}
\end{table}

The existing implementation of modules in Flowcean are summarized in \Cref{tab:modules}.
Offline environments provide loaders for typical data formats in the CPS domain, like ROS bags or CSV files.
Systems based on ordinary differential equations (ODEs) can be simulated incrementally with the ODE environment.
Transform modules include basic transformations of tabular data, such as feature selection or exploding nested columns of multiple data traces into separate columns.
A sliding window transform provides extraction of slices of bounded history from time series data.
Additionally, we provide adaptive transforms that depend on data, like standardization of data with respect to mean and standard deviation.
For the learner module, we integrate existing libraries such as PyTorch, scikit-learn, and LearnLib as well as their respective model representations.
External libraries written in different programming languages, such as LearnLib written in Java, are integrated via the Remote Procedure Call (RPC) framework gRPC\@.
In terms of metrics, Flowcean offers classical metrics for regression and classification tasks.

Furthermore, Flowcean comes with a set of example scenarios to demonstrate the capabilities of the framework.
These scenarios include diverse learning tasks, data sources, and applications.
The subsequent section demonstrates the usage of Flowcean through a case study featuring one of these examples.

\section{Case Study}\label{sec:case_study}

\begin{figure}[htbp]
	\centering
	\tikzset{
  every picture/.style={line width=0.4mm},
}
\begin{tikzpicture}[scale=1]

    \definecolor{myblue}{rgb}{0.1, 0.3, 1.0}
    
    \fill[myblue, fill opacity=0.3] (0,0) -- (0,2.2) -- (4,2.2) -- (4,0.5) -- (5,0.5) -- (5,0) -- cycle;

    \draw (5,0) -- (0,0) -- (0,3);
    \draw (4,3) -- (4,0.5) -- (5,0.5);

    \fill[myblue, fill opacity=0.3] (-1,4.2) -- (2,4.2) -- (2,3) -- (1.3,3) -- (1.3,3.5) -- (-1,3.5) -- cycle;

    \draw (-1,4.2) -- (2,4.2) -- (2,3);
    \draw (-1,3.5) -- (1.3,3.5) -- (1.3,3);

    \draw[->] (1.65,3.5) -- (1.65,2.5);
    \node at (2.4, 2.7) {$b V(t)$};

    \draw[|-|] (-0.5,0) -- (-0.5,2.3);
    \node at (-1, 1.15) {$x$};

    \draw[|-] (0.02,1.6) -- (1.8,1.6);
    \draw[-|] (2.2,1.6) -- (3.98,1.6);
    \node at (2, 1.6) {$A$};

    \draw[->] (4,0.25) -- (5.5,0.25);
    \node at (5.5, 0.6) {$a \sqrt{x}$};

\end{tikzpicture}
	\caption{Non-linear water tank system with inflow $b V(t)$, water level $x$ and resulting outflow $a \sqrt{x}$}\label{fig:tank}
\end{figure}
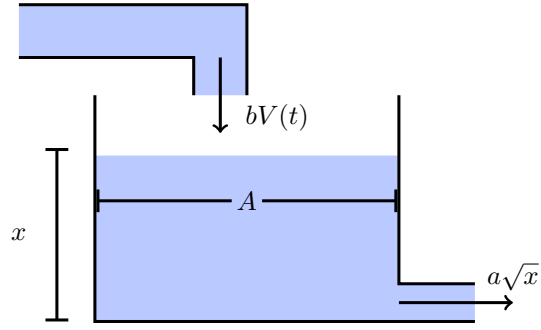

The following case study serves as an illustration of the capabilities of Flowcean, utilizing a well-known one-tank example~\cite{mathworksWatertankSimulinkModel2024}.
It is essential to note that this case study is presented solely to demonstrate the functionality and versatility of the interfaces and abstractions provided by Flowcean.
We acknowledge that more complex CPS scenarios exist; however, the example serves as a suitable test bed to showcase the fundamental principles and functionalities.
The system comprises a tank with a level-dependent outflow and variable inflow, as depicted in \Cref{fig:tank}.
The dynamics of the system are described by the following differential equation:

\[
	\dot{x} = \frac{bV(t) - a\sqrt{x}}{A}
\]\label{eq:ode}
with
\[
	V(t) = \max\left(0,\sin\left(\frac{2\pi}{10}t\right)\right)
\]
where \(x\) denotes the current fill level of the tank, \(t\) the continuous time, \(V(t)\) the time-dependent inflow, \(A\) the area of the tank, and \(a\) and \(b\) the scaling coefficients of the equation.
The initial condition for the simulation is chosen as \(x(0) = 1\).
Furthermore, the system is parameterized with \(A = 5\), \(a = 0.5\), and \(b = 2\).

\begin{figure}[htbp]
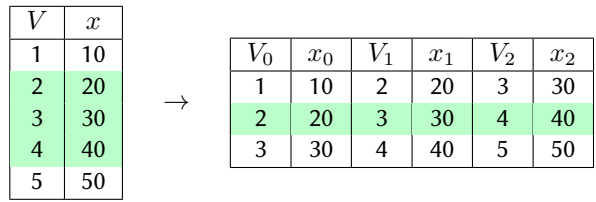

	\centering
\definecolor{highlightcolor}{rgb}{0.1, 1.0, 0.3}

\begin{tabular}{|c|c|}
\hline
\textbf{$V$} & \textbf{$x$} \\ \hline
1 & 10 \\
\cellcolor{highlightcolor!30}2 & \cellcolor{highlightcolor!30}20 \\
\cellcolor{highlightcolor!30}3 & \cellcolor{highlightcolor!30}30 \\
\cellcolor{highlightcolor!30}4 & \cellcolor{highlightcolor!30}40 \\
5 & 50 \\ \hline
\end{tabular}
\hspace{1em} 
\textbf{$\rightarrow$}
\hspace{1em} 
\begin{tabular}{|c|c|c|c|c|c|} \hline
\textbf{$V_0$} & \textbf{$x_0$} & \textbf{$V_1$} & \textbf{$x_1$} & \textbf{$V_2$} & \textbf{$x_2$} \\ \hline
1 & 10 & 2 & 20 & 3 & 30 \\
\cellcolor{highlightcolor!30}2 & \cellcolor{highlightcolor!30}20 & \cellcolor{highlightcolor!30}3 & \cellcolor{highlightcolor!30}30 & \cellcolor{highlightcolor!30}4 & \cellcolor{highlightcolor!30}40 \\
3 & 30 & 4 & 40 & 5 & 50 \\ \hline
\end{tabular}
	\caption{Pivot time series of $V$ and $x$ to extract a trace of bounded history with 3 time steps}\label{fig:sliding_window}
\end{figure}

We implement the system behavior within Flowcean using an \texttt{OdeEnvironment}, which allows for sampling a simulated system described by an ODE\@.
A data set of 250 samples is collected from the simulation at a sampling rate of $0.1 \si{\second}$.
Each sample \(i\) consists of the current inflow \(V_{i}\) and the resulting level \(x_{i}\).
To predict the subsequent fill level \(x_{i + 1}\) based on the current inflow \(V_{i + 1}\) and the two preceding samples \(V_{i}\) and \(V_{i - 1}\), the data set undergoes a \texttt{SlidingWindow} transform whose general function is shown in \Cref{fig:sliding_window}.
The resulting data set, consisting of 248 six-element samples, is split into a training and an evaluation environment where 80\% of the data is used for training and 20\% is used for the evaluation.

An offline learning strategy with two different learners from two different libraries is used to learn models from the training set:

\begin{enumerate}
	\item A regression tree using the scikit-learn library and
	\item A multi-layer perceptron using Lightning, a high-level wrapper for PyTorch.
\end{enumerate}

\begin{figure}[htbp]
	\centering
	\tikzset{
  every picture/.style={line width=0.4mm},
}
\usetikzlibrary{shadings}
\begin{tikzpicture}[every node/.style={align=center}]

    \definecolor{mygreen}{rgb}{0.3, 0.3, 0.3}
    \definecolor{myblue}{rgb}{0.2, 0.3, 1.0}
    \definecolor{myred}{rgb}{0.8, 0.1, 0.1}
    \definecolor{incrementalcolor}{rgb}{1.0, 0.3, 0.8}

    \tikzstyle{box} = [draw, rounded corners=2pt, minimum width=2.5cm, minimum height=1cm, fill opacity=0.3, text opacity=1]
    
    \node[box, fill=mygreen] (ode) at (0.1, 0) {ODE Simulation};
    \node[box, fill=mygreen] (dataset) at (2.9, 0) {Offline Dataset\\
    + Transforms
    };
    \node[box, fill=myblue]  (pytorch) at (5.9, 0.7) {PyTorch};
    \node[box, fill=myblue]  (sklearn) at (5.9, -0.7) {scikit-learn};
    \node[box, fill=myred]  (nn) at (9.0, 0.7) {Neural Network};
    \node[box, fill=myred]  (dt) at (9.0, -0.7) {Decision Tree};
    \node[box, fill=incrementalcolor]  (mae) at (12.4, -0.7) {Mean Absolute Error};
    \node[box, fill=incrementalcolor]  (mse) at (12.4, 0.7) {Mean Squared Error};

    \draw[->] (ode) -- (dataset);
    \draw[->] (dataset) -- (pytorch.west);
    \draw[->] (dataset) -- (sklearn.west);
    \draw[->] (pytorch) -- (nn);
    \draw[->] (sklearn) -- (dt);
    \draw[->] (dt) -- (mae);
    \draw[->] (dt) -- (mse);
    \draw[->] (nn) -- (mae);
    \draw[->] (nn) -- (mse);

    \draw[dashed, rounded corners=2pt] (-1.4, -1.5) rectangle (4.3, 1.5);
    \node at (1.5, 1.8) {Environment};
    \draw[dashed, rounded corners=2pt] (4.4, -1.5) rectangle (7.4, 1.5);
    \node at (5.9, 1.8) {Learner};
    \draw[dashed, rounded corners=2pt] (7.5, -1.5) rectangle (10.5, 1.5);
    \node at (9.0, 1.8) {Model};
    \draw[dashed, rounded corners=2pt] (10.6, -1.5) rectangle (14.2, 1.5);
    \node at (12.4, 1.8) {Metric};

\end{tikzpicture}
	\caption{Modules of the case study implemented in Flowcean composing the learning and evaluation pipeline.}\label{fig:case-study}
\end{figure}
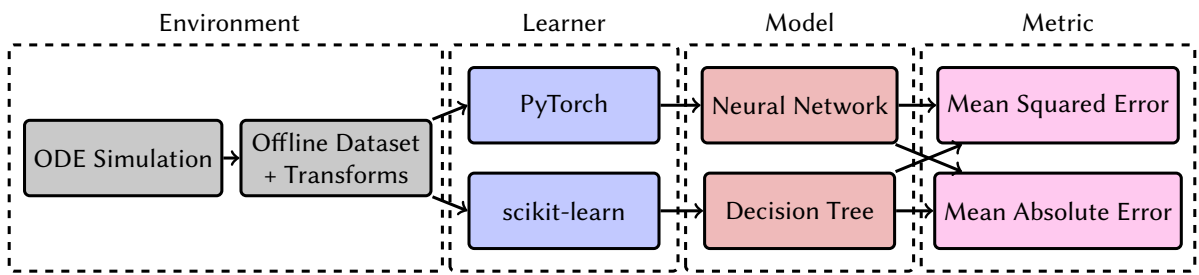

The Mean Absolute Error (MAE) and the Mean Square Error (MSE) on the evaluation environment are computed for both models.
The entire pipeline composed of the modules of \Cref{fig:overview} is visualized in \Cref{fig:case-study}.

These metrics, along with the respective training time, are shown in \Cref{tab:performance}.
For this case study, the regression tree performs significantly better than the multi-layer perceptron.
However, the numerical results of this case study are presented solely to illustrate the workflow of Flowcean.

\begin{table}[htbp]
	\centering
	\caption{Predictive performance of learned models}\label{tab:performance}
	\begin{tabular}{cccc}
		                       & \textbf{Runtime [ms]} & \textbf{MAE} & \textbf{MSE} \\
		\midrule
		Regression Tree        & $15.5$                & $0.0206$     & $0.0006$     \\
		Multi-layer perceptron & $813$                 & $0.0639$     & $0.0054$     \\
	\end{tabular}
\end{table}

The specification of the pipeline for this example using the Flowcean API takes less than one hundred lines of Python code, including all imports and comments.
To create a model using a different learner, only a single line of code needs to be modified, as shown in \Cref{lst:code}.

\begin{figure}[htbp]
	\centering
	\begin{lstlisting}[language=Python, caption={Definition of learners using the Flowcean API}, label={lst:code}]
tree_learner = RegressionTree(max_depth=5)
neural_learner = LightningLearner(max_epochs=100)
\end{lstlisting}
\end{figure}

All other code can be reused without any changes.
Consequently, Flowcean's modular design facilitates the creation of compact and reusable machine learning pipelines.
The complete code for this case study can be accessed on GitHub\footnote{\url{https://github.com/flowcean/flowcean}}.

\section{Conclusion \& Future Work}\label{sec:conclusion}

Understanding and modeling the behavior of CPS is essential in various industrial domains.
However, CPS from different applications exhibit highly diverse compositions of components, which require specific domain knowledge and a profound understanding of the underlying data and learning strategies for modeling.
Constructing pipelines for training and deploying machine learning models tailored to specific problems is a complex and time-consuming challenge.
To address this challenge, we introduce Flowcean, a modular learning framework that addresses the automatic modeling of CPS\@.
Flowcean encompasses common machine learning approaches in a modular architecture.
By this, the framework seamlessly integrates various existing machine learning libraries to allow the configuration of individual modeling pipelines.
Trained models can be evaluated using common metrics and compared to choose suitable models for a given application.

Future work focuses on integrating learned models in application modules for CPS\@.
For this purpose, Flowcean is expanded to include monitoring agents and test case generators as application concepts at the end of the learning workflow.
In addition, real-world examples for the different learning strategies are included in the examples of Flowcean representing applications in energy networks, port technologies, and logistics robots.

\section{Acknowledgement}

This work is funded by BMBF project AGenC no. 01IS22047A.

\bibliography{references}

@article{cuiAnalogCircuitsFault2016,
  title        = {Analog Circuits Fault Diagnosis Using Multi-valued {{Fisher}}'s Fuzzy Decision Tree ({{MFFDT}})},
  author       = {Cui, Yiqian and Shi, Junyou and Wang, Zili},
  year         = 2016,
  month        = jan,
  journal      = {International Journal of Circuit Theory and Applications},
  volume       = 44,
  number       = 1,
  pages        = {240--260},
  doi          = {10.1002/cta.2075},
  issn         = {0098-9886, 1097-007X},
  urldate      = {2024-04-26},
  copyright    = {Copyright {\copyright} 2015 John Wiley \& Sons, Ltd.},
  langid       = {english}
}

@article{derlerModelingCyberPhysical2012a,
  title        = {Modeling {{Cyber}}--{{Physical Systems}}},
  author       = {Derler, P. and Lee, E. A. and Vincentelli, A. S.},
  year         = 2012,
  month        = jan,
  journal      = {Proceedings of the IEEE},
  volume       = 100,
  number       = 1,
  pages        = {13--28},
  doi          = {10.1109/JPROC.2011.2160929},
  issn         = {0018-9219, 1558-2256},
  urldate      = {2024-04-26},
  copyright    = {https://ieeexplore.ieee.org/Xplorehelp/downloads/license-information/IEEE.html},
  langid       = {english}
}

@article{fanReviewDataPreprocessing2021,
  title        = {A {{Review}} on {{Data Preprocessing Techniques Toward Efficient}} and {{Reliable Knowledge Discovery From Building Operational Data}}},
  author       = {Fan, Cheng and Chen, Meiling and Wang, Xinghua and Wang, Jiayuan and Huang, Bufu},
  year         = 2021,
  month        = mar,
  journal      = {Frontiers in Energy Research},
  publisher    = {Frontiers},
  volume       = 9,
  pages        = 652801,
  doi          = {10.3389/fenrg.2021.652801},
  issn         = {2296-598X},
  urldate      = {2024-04-26},
  langid       = {english}
}

@inproceedings{fengTimeSeriesAnomaly2021,
  title        = {Time {{Series Anomaly Detection}} for {{Cyber-physical Systems}} via {{Neural System Identification}} and {{Bayesian Filtering}}},
  author       = {Feng, Cheng and Tian, Pengwei},
  year         = 2021,
  month        = aug,
  booktitle    = {Proceedings of the 27th {{ACM SIGKDD Conference}} on {{Knowledge Discovery}} \& {{Data Mining}}},
  publisher    = {Association for Computing Machinery},
  address      = {Virtual Event Singapore},
  series       = {{{KDD}} '21},
  pages        = {2858--2867},
  doi          = {10.1145/3447548.3467137},
  isbn         = {978-1-4503-8332-5},
  urldate      = {2024-04-26},
  langid       = {english}
}

@inproceedings{khanapuriLearningBasedAdversarialAgent2019,
  title        = {Learning-{{Based Adversarial Agent Detection}} and {{Identification}} in {{Cyber Physical Systems Applied}} to {{Autonomous Vehicular Platoon}}},
  author       = {Khanapuri, Eshaan and Chintalapati, Tarun and Sharma, Rajnikant and Gerdes, Ryan},
  year         = 2019,
  month        = may,
  booktitle    = {2019 {{IEEE}}/{{ACM}} 5th {{International Workshop}} on {{Software Engineering}} for {{Smart Cyber-Physical Systems}} ({{SEsCPS}})},
  publisher    = {IEEE},
  address      = {Montreal, QC, Canada},
  pages        = {39--45},
  doi          = {10.1109/SEsCPS.2019.00014},
  isbn         = {978-1-72812-282-3},
  urldate      = {2024-04-26},
  copyright    = {https://ieeexplore.ieee.org/Xplorehelp/downloads/license-information/IEEE.html},
  langid       = {english}
}

@inproceedings{knittAutomaticGenerationModels2023,
  title        = {Towards the {{Automatic Generation}} of {{Models}} for {{Prediction}}, {{Monitoring}}, and {{Testing}} of {{Cyber-Physical Systems}}},
  author       = {Knitt, Markus and Plambeck, Swantje and Wieck, Jan Christian and Kohlisch, Julian and Balduin, Stephan and Veith, Eric Msp and Schyga, Jakob and Hinckeldeyn, Johannes and Fey, Goerschwin and Kreutzfeldt, Jochen},
  year         = 2023,
  month        = sep,
  booktitle    = {2023 {{IEEE}} 28th {{International Conference}} on {{Emerging Technologies}} and {{Factory Automation}} ({{ETFA}})},
  publisher    = {IEEE},
  address      = {Sinaia, Romania},
  pages        = {1--4},
  doi          = {10.1109/ETFA54631.2023.10275706},
  isbn         = 9798350339918,
  urldate      = {2024-04-26},
  copyright    = {https://doi.org/10.15223/policy-029}
}

@article{kumarsharmaDataDrivenPredictive2022,
  title        = {Data Driven Predictive Maintenance Applications for Industrial Systems with Temporal Convolutional Networks},
  author       = {Kumar Sharma, Deepak and Brahmachari, Shikha and Singhal, Kartik and Gupta, Deepak},
  year         = 2022,
  month        = jul,
  journal      = {Computers \& Industrial Engineering},
  volume       = 169,
  pages        = 108213,
  doi          = {10.1016/j.cie.2022.108213},
  issn         = {03608352},
  urldate      = {2024-09-09},
  langid       = {english}
}

@inproceedings{maierOnlinePassiveLearning2014,
  title        = {Online Passive Learning of Timed Automata for Cyber-Physical Production Systems},
  author       = {Maier, Alexander},
  year         = 2014,
  month        = jul,
  booktitle    = {2014 12th {{IEEE International Conference}} on {{Industrial Informatics}} ({{INDIN}})},
  publisher    = {IEEE},
  address      = {Porto Alegre RS, Brazil},
  pages        = {60--66},
  doi          = {10.1109/INDIN.2014.6945484},
  isbn         = {978-1-4799-4905-2},
  urldate      = {2024-04-26},
  langid       = {english}
}

@misc{mathworksWatertankSimulinkModel2024,
  title        = {Watertank {{Simulink Model}}},
  author       = {{Mathworks}},
  year         = 2024,
  journal      = {watertank Simulink Model},
  urldate      = {2024-04-26},
  howpublished = {https://de.mathworks.com/help/slcontrol/ug/watertank-simulink-model.html}
}

@article{meleshkoMachineLearningBased2020,
  title        = {Machine Learning Based Approach to Detection of Anomalous Data from Sensors in Cyber-Physical Water Supply Systems},
  author       = {Meleshko, A V and Desnitsky, V A and Kotenko, I V},
  year         = 2020,
  month        = jan,
  journal      = {IOP Conference Series: Materials Science and Engineering},
  volume       = 709,
  number       = 3,
  pages        = {033034},
  doi          = {10.1088/1757-899X/709/3/033034},
  issn         = {1757-8981, 1757-899X},
  urldate      = {2024-09-09},
  langid       = {english}
}

@book{ofjallOnlineLearningRobot2014,
  title        = {Online {{Learning}} for {{Robot Vision}}},
  author       = {{\"O}fj{\"a}ll, Kristoffer},
  year         = 2014,
  month        = oct,
  publisher    = {Link{\"o}ping University Electronic Press},
  doi          = {10.3384/lic.diva-110892},
  urldate      = {2024-04-26},
  langid       = {english}
}

@inproceedings{plambeckViabilityDecisionTrees2022,
  title        = {On the {{Viability}} of {{Decision Trees}} for {{Learning Models}} of {{Systems}}},
  author       = {Plambeck, Swantje and Schammer, Lutz and Fey, Gorschwin},
  year         = 2022,
  month        = jan,
  booktitle    = {2022 27th {{Asia}} and {{South Pacific Design Automation Conference}} ({{ASP-DAC}})},
  publisher    = {IEEE},
  address      = {Taipei, Taiwan},
  pages        = {696--701},
  doi          = {10.1109/ASP-DAC52403.2022.9712579},
  isbn         = {978-1-66542-135-5},
  urldate      = {2024-04-26},
  copyright    = {https://doi.org/10.15223/policy-029},
  langid       = {english}
}

@incollection{steffenIntroductionActiveAutomata2011a,
  title        = {Introduction to {{Active Automata Learning}} from a {{Practical Perspective}}},
  author       = {Steffen, Bernhard and Howar, Falk and Merten, Maik},
  year         = 2011,
  booktitle    = {Formal {{Methods}} for {{Eternal Networked Software Systems}}},
  publisher    = {Springer},
  address      = {Berlin, Heidelberg},
  volume       = 6659,
  pages        = {256--296},
  doi          = {10.1007/978-3-642-21455-4_8},
  isbn         = {978-3-642-21455-4},
  urldate      = {2024-04-26},
  editor       = {Bernardo, Marco and Issarny, Val{\'e}rie},
  langid       = {english}
}

@inproceedings{urbatAutomataLearningAlgebraic2020a,
  title        = {Automata {{Learning}}: {{An Algebraic Approach}}},
  shorttitle   = {Automata {{Learning}}},
  author       = {Urbat, Henning and Schr{\"o}der, Lutz},
  year         = 2020,
  month        = jul,
  booktitle    = {Proceedings of the 35th {{Annual ACM}}/{{IEEE Symposium}} on {{Logic}} in {{Computer Science}}},
  publisher    = {ACM},
  address      = {Saarbr{\"u}cken Germany},
  series       = {{{LICS}} '20},
  pages        = {900--914},
  doi          = {10.1145/3373718.3394775},
  isbn         = {978-1-4503-7104-9},
  urldate      = {2024-04-26},
  langid       = {english}
}

@inproceedings{xuDigitalTwinbasedAnomaly2021,
  title        = {Digital {{Twin-based Anomaly Detection}} in {{Cyber-physical Systems}}},
  author       = {Xu, Qinghua and Ali, Shaukat and Yue, Tao},
  year         = 2021,
  month        = apr,
  booktitle    = {2021 14th {{IEEE Conference}} on {{Software Testing}}, {{Verification}} and {{Validation}} ({{ICST}})},
  publisher    = {IEEE},
  address      = {Porto de Galinhas, Brazil},
  pages        = {205--216},
  doi          = {10.1109/ICST49551.2021.00031},
  isbn         = {978-1-72816-836-4},
  urldate      = {2024-04-26},
  copyright    = {https://ieeexplore.ieee.org/Xplorehelp/downloads/license-information/IEEE.html},
  langid       = {english}
}

@article{yaacoubCyberphysicalSystemsSecurity2020,
  title        = {Cyber-Physical Systems Security: {{Limitations}}, Issues and Future Trends},
  shorttitle   = {Cyber-Physical Systems Security},
  author       = {Yaacoub, Jean-Paul A. and Salman, Ola and Noura, Hassan N. and Kaaniche, Nesrine and Chehab, Ali and Malli, Mohamad},
  year         = 2020,
  month        = sep,
  journal      = {Microprocessors and Microsystems},
  volume       = 77,
  pages        = 103201,
  doi          = {10.1016/j.micpro.2020.103201},
  issn         = {0141-9331},
  urldate      = {2024-04-26}
}

@article{yohanandhanCyberPhysicalPowerSystem2020,
  title        = {Cyber-{{Physical Power System}} ({{CPPS}}): {{A Review}} on {{Modeling}}, {{Simulation}}, and {{Analysis With Cyber Security Applications}}},
  shorttitle   = {Cyber-{{Physical Power System}} ({{CPPS}})},
  author       = {Yohanandhan, Rajaa Vikhram and Elavarasan, Rajvikram Madurai and Manoharan, Premkumar and {Mihet-Popa}, Lucian},
  year         = 2020,
  journal      = {IEEE Access},
  volume       = 8,
  pages        = {151019--151064},
  doi          = {10.1109/ACCESS.2020.3016826},
  issn         = {2169-3536},
  urldate      = {2024-04-26},
  copyright    = {https://creativecommons.org/licenses/by/4.0/legalcode},
  langid       = {english}
}

\appendix

\end{document}